\pdfoutput=1

\documentclass[11pt]{article}

\usepackage[final]{coling}
\usepackage{graphicx}
\usepackage{latexsym}
\usepackage{supertabular}
\usepackage{stfloats}
\usepackage{tabularx}
\usepackage{multirow}
\usepackage{makecell}
\usepackage{booktabs}
\usepackage{amsmath}
\usepackage{amsfonts}
\usepackage{amssymb}
\usepackage{enumerate}
\usepackage{algorithm}
\usepackage{algorithmic}
\usepackage{bbm}
\usepackage[dvipsnames]{xcolor}
\usepackage{pifont}
\usepackage{xspace}
\usepackage{inconsolata}
\usepackage{subfigure}
\usepackage{caption}
\usepackage{subcaption}

\usepackage{times}
\usepackage{latexsym}
\usepackage{graphicx}  
\usepackage{float}  
\usepackage{subfigure}  

\usepackage[T1]{fontenc}

\usepackage[utf8]{inputenc}

\usepackage{microtype}

\usepackage{inconsolata}

\usepackage{graphicx}

\title{Multilingual Knowledge Editing with Language-Agnostic Factual Neurons}

\author{Xue Zhang\textsuperscript{1}\thanks{ \ \ This work was done during internship at Pattern Recognition Center, WeChat AI, Tencent Inc, China.}, Yunlong Liang\textsuperscript{2}, 
Fandong Meng\textsuperscript{2}, Songming           Zhang\textsuperscript{1},\\
\textbf{Yufeng Chen}\textsuperscript{1}\thanks{ \ \ Yufeng Chen is the corresponding author.},
\textbf{Jinan Xu}\textsuperscript{1}, \textbf{Jie Zhou}\textsuperscript{2} \\
\textsuperscript{1}Beijing Key Lab of Traffic Data Analysis and Mining, \\
Beijing Jiaotong University, Beijing, China \\
\textsuperscript{2}Pattern Recognition Center, WeChat AI, Tencent Inc, China \\
\texttt{\{\text{zhang\_xue},smzhang22,chenyf,jaxu\}@bjtu.edu.cn}, \\
\texttt{\{yunlonliang,fandongmeng,withtomzhou\}@tencent.com} \\}

\begin{document}
\maketitle
\begin{abstract}

Multilingual knowledge editing (MKE) aims to simultaneously update factual knowledge across multiple languages within large language models (LLMs).
Previous research indicates that the same knowledge across different languages within LLMs exhibits a degree of shareability.
However, most existing MKE methods overlook the connections of the same knowledge between different languages, resulting in knowledge conflicts and limited edit performance.
To address this issue, we first investigate how LLMs process multilingual factual knowledge and discover that the same factual knowledge in different languages generally activates a shared set of neurons, which we call language-agnostic factual neurons (LAFNs).
These neurons represent the same factual knowledge shared across languages and imply the semantic connections among multilingual knowledge.
Inspired by this finding, we propose a new MKE method by Locating and Updating Language-Agnostic Factual Neurons (LU-LAFNs) to edit multilingual knowledge simultaneously, which avoids knowledge conflicts and thus improves edit performance.
Experimental results on Bi-ZsRE and MzsRE benchmarks demonstrate that our method achieves the best edit performance, indicating the effectiveness and importance of modeling the semantic connections among multilingual knowledge.

\end{abstract}

\section{Introduction}


Multilingual knowledge editing (MKE) \cite{wang2023retrievalaugmented} aims to simultaneously rectify factual knowledge across multiple languages within large language models (LLMs).
This process poses more challenges \cite{wang2023crosslingual} compared to monolingual knowledge editing (KE) since the edited factual knowledge should be updated together across multiple languages.

\begin{table}[t]
    \centering
    \resizebox*{\linewidth}{!}{
    \begin{tabular}{c|cc|ll}
    \bottomrule     
        
        \textbf{Edit Languages} &  \multicolumn{2}{c|}{\textbf{Monolingual}}	&	\multicolumn{2}{c}{\textbf{Multilingual}} \\
        
        \hline
        \textbf{Test Languages} &  \textbf{en} &	\textbf{zh} &	\makecell[c]{\textbf{en}}  &	\makecell[c]{\textbf{zh}} \\
        
         \hline
LoRA-FT	&	48.46  &	34.33 	&	47.06 \color{Red}{($-${1.4})}		&	33.17 \color{Red}{($-${1.16}})	\\
M-ROME	&	65.84  &	53.66 	&	54.82 \color{Red}{($-$\textbf{11.02}}) 		&	46.27 \color{Red}{($-$\textbf{7.40}})	\\
M-MEMIT	&	68.18  &	62.80 	&	58.63 \color{Red}{($-$9.55})		&	57.16  \color{Red}{($-$5.64})	\\
M-PMET	&	68.18  &	60.97 	&	61.98 \color{Red}{($-$6.19}) 		&	57.33 \color{Red}{($-$3.64}) 	\\
\textbf{LU-LAFNs (Ours)}	&	\textbf{72.85}  &	 \textbf{66.71}	&	\textbf{73.92} \color{ForestGreen}{($+$\textbf{1.07}})  		&	\textbf{67.29} \color{ForestGreen}{($+$\textbf{0.58}}) 	\\

    \toprule
    \end{tabular}
    }
    \caption{
        The average EM results of Reliability, Generality, Locality, and Portability on Bi-ZsRE using Llama-3.1-8B as the backbone. 
        ``Monolingual'' means editing and testing on the same one language, while ``Multilingual'' means editing on both $en$ and $zh$, and testing on each language, respectively.
        The values in {\color{Red}red} represent the performance decline compared to monolingual KE due to knowledge conflicts across languages.  
        The values in {\color{ForestGreen}green} indicate that our method avoids such conflicts and further promotes the edit performance compared to monolingual KE. 
    }
    \label{table:introduction}
\end{table}

Recently, some Locate-then-Edit \cite{yao-etal-2023-editing} methods, such as ROME \cite{NEURIPS2022_6f1d43d5}, MEMIT \cite{meng2023massediting}, and PMET \cite{li2024pmet}, exhibit strong edit performance in monolingual KE.
These methods identify parameters corresponding to specific knowledge and directly modify them to the target parameters.
When adapting them to MKE, edit performance will probably degrade due to the conflicts between different languages (as shown in the red results of Table \ref{table:introduction}).
Similarly, directly fine-tuning the original model with LoRA \cite{hu2021lora} also suffers from performance degradation due to the multilingual knowledge conflicts in LoRA modules.
These conflicts can be attributed to the ignoring of the potential connections between multilingual knowledge in LLMs \cite{chen2023journey}.
Therefore, it is important to model the connections between multilingual knowledge during the editing process to avoid such conflicts.

To address this problem, we first investigate how LLMs process the same factual knowledge in different languages.
We discover that the same multilingual factual knowledge generally activates a shared set of neurons in feed-forward networks (FFNs), which we call Language-Agnostic Factual Neurons (LAFNs). 
These neurons represent the same factual knowledge shared across multiple languages and imply the semantic connections among multilingual knowledge.
Based on this finding, we propose a new MKE method by \textbf{L}ocating and \textbf{U}pdating \textbf{L}anguage-\textbf{A}gnostic \textbf{F}actual \textbf{N}eurons (\textbf{LU-LAFNs}) to edit multilingual knowledge simultaneously.
Specifically, 
we generate a set of paraphrases for multilingual knowledge to precisely locate LAFNs.
Then we optimize the update values for modifying these located neurons to achieve simultaneous modification of the same multilingual knowledge.
Additionally, to avoid the degradation of the edited model's general abilities due to directly modifying model parameters \cite{gu2024model}, we store the update values of the edited LAFNs in the cache.
When the edited subject appears in the user query, the relative update values will be retrieved and used for model inference.

To evaluate the effectiveness of our method, we conduct experiments on two multilingual KE benchmarks, Bi-ZsRE \cite{wang2023crosslingual} and MzsRE \cite{wang2023retrievalaugmented}.
Experimental results demonstrate that our method outperforms existing MKE methods in terms of Reliability, Generality, and Locality.
Further analysis indicates that our method avoids conflicts by modeling the semantic connections between multilingual knowledge and thus improves the edit performance.

In summary, the major contributions of this paper are as follows\footnote{The code is publicly available at \url{https://github.com/XZhang00/LU-LAFNs}.}:
\begin{itemize}
    \item We propose a new method by locating and updating language-agnostic factual neurons to achieve MKE. Our method avoids conflicts by modeling the semantic connections between multilingual knowledge.
    \item Experimental results on Bi-ZsRE and MzsRE benchmarks demonstrate that our method achieves the best edit performance, which proves the effectiveness of our method. 
    \item 
    We further analyze the key factors that influence multilingual edit performance, including LLMs' inherent language capabilities, the updated layers, and the number of LAFNs.
\end{itemize}

\section{Related Work}

\noindent\textbf{Multilingual Knowledge Editing.} 
MKE aims to update multilingual knowledge simultaneously by using parallel multilingual data.
ReMaKE \cite{wang2023retrievalaugmented} retrieves the multilingual aligned knowledge from a multilingual knowledge base as context to achieve MKE.
Additionally, some methods, such as LiME \cite{xu-etal-2023-language-anisotropic} and MPN \cite{si2024mpn}, explore cross-lingual knowledge editing, which only utilizes monolingual knowledge to edit the model and then test the edit performance on other languages.
In this work, we mainly focus on MKE, which is more practical and performs better in updating multilingual outdated knowledge.

\noindent\textbf{Multilingual Knowledge Analysis.}
Analyzing the multilingual capabilities of language models is always a research hotspot \cite{pires-etal-2019-multilingual, bhattacharya-bojar-2023-unveiling, kojima2024multilingual, zhao2024large}, especially exploring the relationship between model architecture and multilingual capabilities.
\citet{tang2024languagespecific} indicate that LLMs’ proficiency in processing a particular language is predominantly due to a subset of neurons within FFNs.
Similar to our work, \citet{chen2023journey} discover the language-independent knowledge neurons of mBERT and mGPT, which store knowledge in a form that transcends language, but ignores how to control neurons to achieve desired outputs.
Differently, we first investigate the language-agnostic factual neurons related to specific fact knowledge in LLMs and then modify them to achieve MKE.

\section{Methodology}
In this section, we first give the definition of MKE (\S\ref{sec:3-0}).
Then we investigate how LLMs process multilingual factual knowledge by identifying and analyzing the associated neurons (\S\ref{sec:3-1}).
Subsequently, we introduce our method LU-LAFNs for MKE (\S\ref{sec:3-2}).

\subsection{Task Definition}\label{sec:3-0}

MKE aims to simultaneously update multilingual knowledge with new information while preserving previous accurate knowledge within the model.
Formally, we denote the original model as $\mathcal{F}_{\theta}$ and the multilingual group of an edit descriptor $(x^e, y^e)$  as ${G}\!=\!\{(x^e_{\ell}, {y}^e_{\ell})|\ell \in {L}\}$,
where $x^{e}_\ell$ is the question for the knowledge to be edited in language $\ell$ and usually contains a subject and a relation, and ${y}^e_{\ell}$ is the new answer of $x^e_\ell$.
On this basis, MKE will lead to a model $\mathcal{F}_{\theta}^{'}$ to correctly answer the edited question $x^{e}_\ell$ in each language $\ell$ and meanwhile maintain the original prediction on other unedited questions:
\begin{equation}
\forall \ell \in L, {\mathcal{F}_{\theta}^{'}(x_{\ell})} \!=\! 
\left\{
\begin{array}{ll}
    y^{e}_{\ell}, & x_{\ell} \in I(x^{e}_{\ell}),  \\
\mathcal{F}_{\theta}(x_{\ell}), & x_{\ell} \notin I(x^{e}_{\ell}),
\end{array}
\right.
\end{equation}
where $I(x^e_{\ell})$ denotes a broad set of inputs with the same semantics as $x^e_{\ell}$ \cite{wang2023crosslingual}.



\subsection{Language-Agnostic Factual Neurons}\label{sec:3-1}

To investigate how LLMs process the same factual knowledge represented in different languages, we identify and analyze language-agnostic factual neurons (LAFNs) within FFNs based on two multilingual LLMs. 
Specifically, we first separately identify the factual neurons associated with monolingual knowledge in each language. 
Then, we take the intersection of these neurons across different languages to obtain the LAFNs.

\paragraph{Identifying LAFNs.} 

The forward process of the FFN layer in current LLMs can generally be described as the following two formulas:
\begin{equation}\label{eq:ffn2}
    h^i = \text{act\_fn}(\Tilde{h}^i W_1^i) \cdot W_2^i, 
\end{equation}
\begin{equation}\label{eq:ffn3}
    h^i = (\text{act\_fn}(\Tilde{h}^i W_1^i) \otimes \Tilde{h}^i  W_2^i) \cdot W_3^i, 
\end{equation}
where $i$ denotes the $i$-th FFN layer,
$\Tilde{h}^i$/$h^i$ are the output hidden states of the $i$-th attention/FFN layer, and $\text{act\_fn}(\cdot)$ is the activation function.
Eq.(\ref{eq:ffn2}) represents the FFN structure of older LLMs, \textit{e.g.}, BLOOM-series \cite{muennighoff2023crosslingualgeneralizationmultitaskfinetuning}.
Eq.3 shows the FFN structure of the latest LLMs, \textit{e.g.}, Qwen2 \cite{qwen2} and Llama3 \cite{dubey2024llama3herdmodels}, where $W_1^i$, $W_2^i$, $W_3^i$ correspond to the \texttt{gate\_proj}, \texttt{up\_proj}, \texttt{down\_proj} matrix, respectively.
In this process, knowledge neurons refer to the output activations by the activation function after the first matrix of FFNs, \emph{i.e.}, ${\text{act\_fn}(\Tilde{h}^i    W_1^i)}$. 
Then we define that the $j$-th neuron of the $i$-th FFN layer 
is activated when ${\text{act\_fn}(\Tilde{h}^i    W_1^i)}_j > 0$ following the previous work \cite{tang2024languagespecific}.

For the factual neurons of language $\ell$,
we use the factual corpus $C_\ell$ in language $\ell$ to track the activation of neurons in each FFN layer during the forward propagation. 
Subsequently, we identify and select the neurons that are activated most frequently to form the neuron set.
For instance, the set of factual neurons in the $i$-th FFN layer $D^i_{\ell}$ can be obtained using $C_{\ell}$ as follows: 
\begin{equation}\label{eq:n1}
    N^i \!=\! \big\{{n^i_j | n^i_j \!=\! \sum_{c \in C_\ell}\!\mathbbm{1}({\text{act\_fn}(\Tilde{h}^i_c W_1^i)}_j \!>\! 0)}\big\},
\vspace{-5pt}
\end{equation}
\begin{equation}\label{eq:n2}
    D^i_{\ell} \!=\! \{ j \mid \frac{n^i_j}{\max(N^i)}  \!>\! \beta\}, 
\end{equation}
where $\Tilde{h}^i_c$ contains $\Tilde{h}^i$ at each token position in sentence $c$, 
$\mathbbm{1}({\text{act\_fn}(\Tilde{h}^i_c W_1^i)}_j > 0)$ equals to $1$ when ${\text{act\_fn}(\Tilde{h}^i_cW_1^i)}_j > 0$ otherwise $0$, $n^i_j$ is the total activation counts of the $j$-th neuron of the $i$-th FFN layer, $N^i$ is the set of activation counts of all neurons in $i$-th FFN layer when processing $C_{\ell}$, and $\beta$ is the threshold to control the amount of $D^i_{\ell}$.
After obtaining the sets of factual neurons for each language in ${L}$, we calculate the intersection of all these sets in the $i$-th FFN layer to extract the shared knowledge neurons among all languages:
\begin{equation}\label{eq:neurons}
D^i = D^i_{\ell_1} \cap D^i_{\ell_2} \cap \dots \cap D^i_{\ell_L},
\end{equation}
where we call $D^i$ as the LAFNs in the $i$-th layer that imply the semantic connections of $\{C_\ell, \ell \in L\}$.


\paragraph{Experiments.} 
We conduct analysis on PARAREL \cite{elazar2021measuring}, which contains factual knowledge with 34 relations in English.
Here, we identify the LAFNs between English ($en$) and Chinese ($zh$).
Firstly, we randomly choose 3000 sentences in each relation from PARAREL to build the factual corpus $C_{en}$ (around 100k), and then utilize the \texttt{Google Translate API} to translate $C_{en}$ to $C_{zh}$.
We select two public multilingual LLMs: Llama-3.1-8B \cite{dubey2024llama3herdmodels} and Qwen2-7B \cite{qwen2}.
The layer numbers of the two LLMs are 32 and 28.
The threshold $\beta$ in Eq.(\ref{eq:n2}) for two LLMs is set to 0.9 and 0.8.
According to Eq.(\ref{eq:n2}) and Eq.(\ref{eq:neurons}), we count the LAFNs in each layer for the two LLMs.

\begin{figure}[t]
    \centering
    \includegraphics[width=\linewidth]{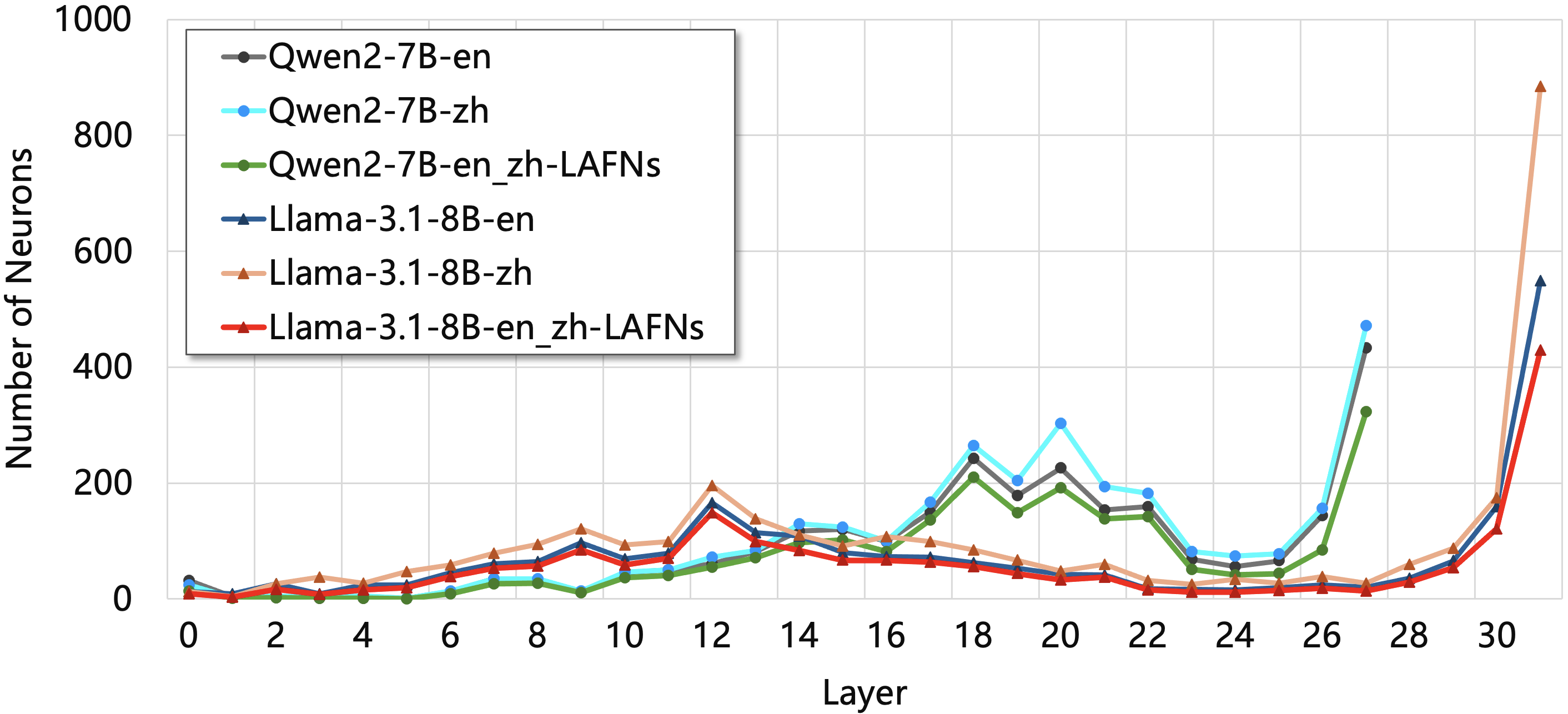}
    \caption{The identified neuron numbers in each layer of Qwen2-7B and Llama-3.1-8B. ``xxx-en'' and ``xxx-zh'' represent the English and Chinese factual neurons respectively. ``xxx-LAFNs'' refers to the language-agnostic factual neurons shared by English and Chinese.}
    \label{fig:layer}
\end{figure}

\paragraph{Results.} 
We plot the identified neuron numbers in each layer of the two LLMs in Figure \ref{fig:layer}, including the factual neurons of each language and LAFNs, \emph{i.e.}, $D_{en}^{i}$, $D_{zh}^{i}$ and $D^{i}$. 
It shows that the changes of the neuron numbers for the two models exhibit similar trends, with a greater presence of language-agnostic knowledge neurons in the middle layers and the last layer (refer to the green and red lines in Figure \ref{fig:layer}). 
The difference is that in the middle layers, Llama-3.1-8B peaks in quantity at the 12th layer, while Qwen2-7B reaches its peak at the 18th layer.
These results prove the existence of LAFNs, which represent the connections between the same factual knowledge in different languages and are mainly located in certain layers.
Based on this finding, we design our method LU-LAFNs.

\begin{figure*}[t]
    \centering
    \includegraphics[width=\linewidth]{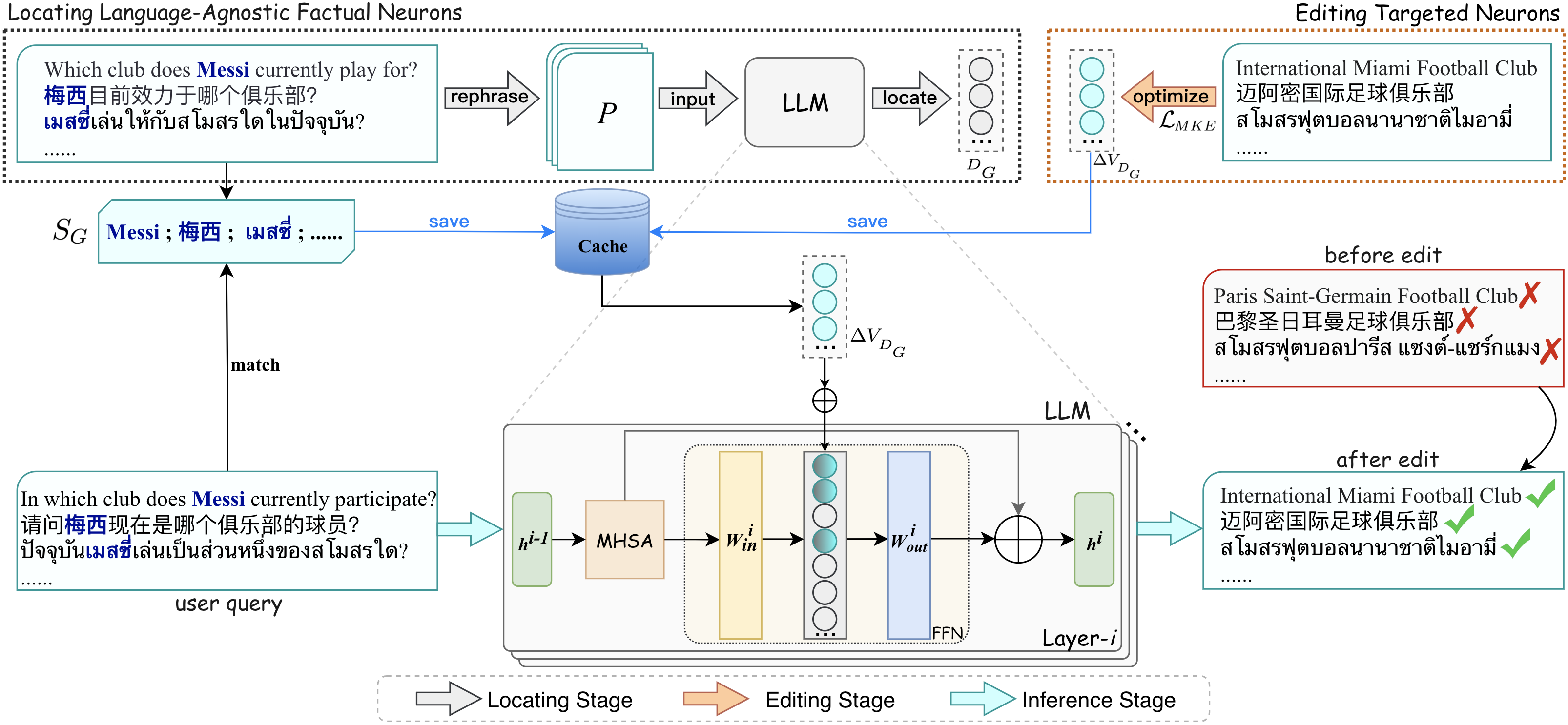}
    \caption{The architecture of our LU-LAFNs. Given the multilingual knowledge to be edited (including the aligned multilingual subject set $S_G$), we first locate the corresponding language-agnostic factual neurons $D_G$. Then the update values $\Delta V_{D_G}$ is optimized for modifying $D_G$, and $\{S_G: \Delta V_{D_G}\}$ is stored in cache. When the subject of the user query is matched in the cache, the relative $\Delta V_{D_G}$ is used for model inference.}
    \label{fig:model}
\end{figure*}

\subsection{LU-LAFNs}\label{sec:3-2}

Figure \ref{fig:model} shows the architecture of our method.
We first locate the LAFNs for each group of multilingual edit descriptors, and then we optimize the update values to modify these neurons and store them in the cache.
During the inference stage, when the subject of the user query is matched in the cache, the relative update values are utilized for model inference.


\paragraph{Locating Stage.}
Given the multilingual group $G$ of an edit descriptor $(x^e, y^e)$ ($G\!=\!\{(x^e_{\ell}, {y}^e_{\ell})|\ell \in L\}$),
we first locate the factual neurons $D_{\ell}^i$ in $i$-th layer for $(x^e_{\ell}, {y}^e_{\ell})$ in language $\ell$ according to Eq.(\ref{eq:n1}) and Eq.(\ref{eq:n2}).
Specifically, to more precisely locate the neurons that are semantically related to $x^e_{\ell}$, we use an LLM to generate several paraphrases for $x^e_{\ell}$ to build its paraphrase set as the factual corpus $C_\ell$ in Eq.(\ref{eq:n1}).
After obtaining $D_{\ell}^i$ in each language $\ell$, we follow Eq.(\ref{eq:neurons}) to obtain the LAFNs set $D^i_G$ of $G$ in $i$-th layer.

\paragraph{Editing Stage.}
Given one multilingual edit description group $G$ and its LAFNs set $D_G$ of located layers, we aim to modify the values of $D_G$ to edit knowledge in $G$ simultaneously.
Following the settings of MEMIT \cite{meng2023massediting} and PMET \cite{li2024pmet}, we modify the values $V_{D_G}$ of $D_G$ at the last token position of the subject in the question $x^e_\ell$.
As for subjects, we obtain the corresponding aligned multilingual subject set $S_G$ from $G$ (refer to $S_{G}$ in Figure \ref{fig:model}). 
Then we will optimize the update values $\Delta V_{D_G}$ for adding to $V_{D_G}$ to achieve MKE.
That is, the model should generate the corresponding new answer $y^e_\ell$ by adding the $\Delta V_{D_G}$:
\begin{equation}
    \mathcal{F}_{(\theta, V_{D_G} + \Delta V_{D_G})}(x^e_{\ell}) = {y}^e_{\ell}.
\end{equation}
To this end, we calculate the $\mathcal{L}_{target}$ to optimize $\Delta V_{D_G}$:
\begin{equation}\label{eq:loss}
\begin{aligned}
\mathcal{L}_{target} \!=\! \frac{1}{|L|M} \sum_{\ell \in L} \sum_{m=1}^{M} \!-\!\log {P}_{\mathcal{F}_{\theta}^{'}}({y}^e_{\ell} \mid {p}_{\ell}^m \!+\! x^e_{\ell} ), \\
\end{aligned}
\end{equation}
where $\ell \in {L}$, ${\mathcal{F}_{\theta}^{'}} = \mathcal{F}_{(\theta, V_{D_G} + \Delta V_{D_G})}$, and $p^m_{\ell}$ represents a randomly generated prefix to improve generalization \cite{meng2023massediting} on $I(x^e_{\ell})$, and $M$ is the total number of prefixes.

Additionally, to ensure that the knowledge under the other relations of $S_G$ is not affected, we also use $\mathcal{L}_{kl}$ to optimize $\Delta V_{D_G}$ similar to MEMIT \cite{meng2023massediting} and PMET \cite{li2024pmet}:
\begin{equation}
    \mathcal{L}_{kl} \!=\! \frac{1}{|L|} \sum_{\ell \in L}^{} \mathrm{KL}\big[{P}_{\mathcal{F}_{\theta}}(y \mid q_{\ell}) \!\mid \mid \!{P}_{\mathcal{F}_{\theta}^{'}}(y \mid q_{\ell}) \big],
\end{equation}
where $q_{\ell}$ has the format of ``\{$s_{\ell}$\} is a'' in language $\ell$, $s_{\ell}$ is the subject in $x_\ell^e$ and $s_{\ell} \in S_G$, and $\mathrm{KL}[\cdot \mid \mid \cdot]$ is the Kullback-Leibler divergence \cite{10.1214/aoms/1177729694}.

In the end, the overall optimized objective $\mathcal{L}_{\rm MKE}$ consists of the above two loss functions:
\begin{equation}\label{eq:loss-all}
    \mathcal{L}_{\rm MKE} = {\lambda}_1 \mathcal{L}_{target} + {\lambda}_2 \mathcal{L}_{kl},
\end{equation}
where $\lambda_1$ and $\lambda_2$ are hyperparameters to control the weight of two loss functions.

After obtaining $\Delta V_{D_G}$, we store $\{S_G: \Delta V_{D_G}\}$ in the cache to avoid directly modifying the model parameters.
When the subject $s_{\ell}$ of the current query $x_\ell$ is matched\footnote{Here, we use the exact-match method.} in $S_G$, we retrieve the corresponding $\Delta V_{D_G}$ for model inference as follows:
\begin{equation}\label{eq:infer}
{\mathcal{F}_{\theta}^{'}}(x_\ell) \!=\! 
\left\{
\begin{array}{ll}
\!\mathcal{F}_{(\theta,V_{D_G} + \Delta V_{D_G})}(x_\ell), \!&\! s_\ell \!\in\! S_G. \\
\!\mathcal{F}_{\theta}(x_\ell), \!&\! s_\ell \!\notin\! S_G.
\end{array}
\right.
\end{equation}


\section{Experiments}

\subsection{Experimental Settings}
\paragraph{Datasets and Metrics.}
We conduct our experiments on Bi-ZsRE \cite{wang2023crosslingual} and MzsRE \cite{wang2023retrievalaugmented}.
Bi-ZsRE covers English ($en$) and Chinese ($zh$) languages, and each language contains 10000/3000/1037 samples for the train/development/test set.
MzsRE includes 12 languages\footnote{English ($en$), Chinese ($zh$), Czech ($cz$), German ($de$), Dutch ($nl$), Spanish ($es$), French ($fr$), Portuguese ($pt$), Russian ($ru$), Thai ($th$), Turkish ($tr$), and Vietnamese ($vi$).}, and each language consists of 10000/743 examples for the train/test set.
Following \citet{wang2023crosslingual}, we calculate the F1/EM value of Reliability, Generality, Locality, and Portability as our evaluation metrics. 
The detailed introduction of metrics is listed in Appendix \ref{sec:appendix-metrics}. 

\paragraph{Backbones.}
In our experiments, we select three public multilingual models as backbones to conduct MKE\footnote{In the initial stage, we conduct cross-lingual experiments on Llama2-7B \cite{touvron2023llama}. We list and discuss these results in Appendix \ref{sec:appendix-cross-lingual}.}, including Llama-3.1-8B \cite{dubey2024llama3herdmodels}, Qwen2-7B \cite{qwen2}, and bloomz-7b1-mt \cite{muennighoff2023crosslingualgeneralizationmultitaskfinetuning}.
Among them, each FFN layer of Llama-3.1-8B and Qwen2-7B follows Eq.(\ref{eq:ffn3}), and bloomz-7b1-mt follows Eq.(\ref{eq:ffn2}).
The detailed supported languages of the three LLMs are introduced in Appendix \ref{sec:appendix-langsfor3}.

\paragraph{Implementation Details.}
When locating LAFNs in \S\ref{sec:3-2}, we utilize the Qwen2-72B-instruct \cite{qwen2} model to generate 30 paraphrases for each $x^e_\ell$.
The detailed instruction is listed in Appendix \ref{sec:appendix-para-instuction}.
The length of each randomly generated prefix $p_{\ell}^m$ in Eq.(\ref{eq:loss}) is set to 5, and the total amount $M$ of prefixes for each language is set to 4.
Additionally, ${\lambda}_1$ in Eq.(\ref{eq:loss-all}) is set to 1, and ${\lambda}_2$ is set to 0.0625 following MEMIT \cite{meng2023massediting}.
For layers to be modified, we set (11, 12, 13, 31) for Llama-3.1-8B, (19, 20, 21, 27) for Qwen2-7B, and (9, 10, 11, 29) for bloomz-7b1-mt respectively.
And the threshold $\beta$ in Eq.(\ref{eq:n2}) is set to 0.1, 0, and 0.2 for Llama-3.1-8B, Qwen2-7B, and bloomz-7b1-mt respectively.

\begin{table*}[t]
    \centering
    \resizebox*{\linewidth}{!}{
    \begin{tabular}{c|cccc|cccc|c}
    \bottomrule
        & \multicolumn{4}{c|}{\textbf{Test Language: en}} & \multicolumn{4}{c|}{\textbf{Test Language: zh}} \\
        \hline
        \textbf{Methods} &  \textbf{Reliability} &	\textbf{Generality} &	\textbf{Locality}  &	\textbf{Portability} & \textbf{Reliability} &	\textbf{Generality}	& \textbf{Locality}	&  \textbf{Portability} &	\textbf{\textit{avg}}\\
        
    \toprule
    \multicolumn{10}{c}{\textbf{Llama-3.1-8B \quad (Edit Languages: en \& zh)}} \\
    \bottomrule
        LoRA-FT	&	97.31 	/	95.47 	&	77.39 	/	62.78 	&	52.52 	/	25.75 	&	32.41 	/	\textbf{4.24} 	&	86.59 	/	74.06 	&	75.56 	/	52.07 	&	27.78 	/	2.41 	&	30.32 	/	\textbf{4.15} 	&	50.05 	\\
ReMaKE	&	43.40 	/	16.30 	&	44.84 	/	17.74 	&	\textbf{100.0} 	/	\textbf{100.0} 	&	34.45 	/	0.96 	&	56.66 	/	20.25 	&	57.00 	/	20.44 	&	\textbf{100.0} 	/	\textbf{100.0} 	&	\textbf{37.26} 	/	0.87 	&	46.89 	\\
M-ROME	&	85.89 	/	75.80 	&	79.09 	/	65.67 	&	88.49 	/	73.67 	&	34.18 	/	4.15 	&	82.52 	/	68.18 	&	79.00 	/	61.62 	&	78.44 	/	51.40 	&	31.79 	/	3.86 	&	60.23 	\\
M-MEMIT	&	88.79 	/	80.42 	&	77.30 	/	61.91 	&	96.27 	/	89.10 	&	35.69 	/	3.09 	&	85.92 	/	75.22 	&	80.22 	/	64.80 	&	95.40 	/	84.67 	&	33.59 	/	3.95 	&	66.02 	\\
M-PMET	&	91.34 	/	83.70 	&	81.70 	/	69.14 	&	96.73 	/	90.94 	&	\textbf{35.87} 	/	4.15 	&	85.39 	/	73.87 	&	81.48 	/	67.02 	&	95.28 	/	84.76 	&	33.29 	/	3.66 	&	67.40 	\\
LU-LAFNs (Ours)	&	\textbf{99.34} 	/	\textbf{98.94} 	&	\textbf{96.03} 	/	\textbf{93.44} 	&	\textbf{100.0} 	/	\textbf{100.0} 	&	29.84 	/	3.28 	&	\textbf{90.71} 	/	\textbf{83.99} 	&	\textbf{88.90} 	/	\textbf{81.20} 	&	\textbf{100.0} 	/	\textbf{100.0} 	&	29.01 	/	3.95 	&	\textbf{74.91} 	\\  
    \toprule
    
    \multicolumn{10}{c}{\textbf{Qwen2-7B \quad (Edit Languages: en \& zh)}} \\
    \bottomrule
        LoRA-FT	&	88.82 	/	80.81 	&	69.93 	/	54.10 	&	49.60 	/	20.54 	&	31.72 	/	\textbf{5.79} 	&	96.17 	/	92.29 	&	83.01 	/	67.89 	&	52.31 	/	21.60 	&	34.94 	/	\textbf{7.71} 	&	53.58 	\\
ReMaKE	&	43.64 	/	16.30 	&	44.75 	/	17.55 	&	\textbf{100.0} 	/	\textbf{100.0} 	&	\textbf{34.47} 	/	1.25 	&	62.04 	/	28.25 	&	63.22 	/	29.80 	&	\textbf{100.0} 	/	\textbf{100.0} 	&	\textbf{39.07} 	/	3.47 	&	48.99 	\\
M-ROME	&	81.89 	/	70.97 	&	74.08 	/	59.59 	&	92.74 	/	83.41 	&	33.11 	/	1.83 	&	88.15 	/	77.05 	&	81.52 	/	65.86 	&	93.41 	/	81.68 	&	33.65 	/	4.15 	&	63.94 	\\
M-MEMIT	&	97.17 	/	94.99 	&	89.00 	/	82.26 	&	94.17 	/	86.60 	&	34.18 	/	2.89 	&	98.56 	/	96.24 	&	93.44 	/	87.17 	&	95.62 	/	87.95 	&	33.94 	/	5.21 	&	73.71 	\\
M-PMET	&	88.13 	/	79.27 	&	77.91 	/	64.32 	&	93.59 	/	85.25 	&	34.10 	/	2.22 	&	90.24 	/	80.42 	&	82.53 	/	67.31 	&	95.25 	/	87.17 	&	33.38 	/	3.95 	&	66.57 	\\
LU-LAFNs (Ours)	&	\textbf{99.45} 	/	\textbf{99.23} 	&	\textbf{95.61} 	/	\textbf{92.29} 	&	\textbf{100.0} 	/	\textbf{100.0} 	&	30.27 	/	2.03 	&	\textbf{99.80} 	/	\textbf{99.71} 	&	\textbf{96.50} 	/	\textbf{93.06} 	&	\textbf{100.0} 	/	\textbf{100.0} 	&	30.78 	/	5.11 	&	\textbf{77.74} 	\\
    \toprule

    \multicolumn{10}{c}{\textbf{bloomz-7b1-mt \quad (Edit Languages: en \& zh)}} \\
    \bottomrule
        LoRA-FT	&	83.76 	/	75.31 	&	64.11 	/	48.60 	&	29.63 	/	7.62 	&	23.14 	/	\textbf{3.66} 	&	94.49 	/	89.39 	&	78.52 	/	64.03 	&	18.21 	/	3.38 	&	22.55 	/	4.15 	&	44.41 	\\
ReMaKE	&	28.78 	/	2.03 	&	28.30 	/	1.16 	&	\textbf{100.0} 	/	\textbf{100.0} 	&	22.29 	/	0.00 	&	61.08 	/	37.99 	&	60.77 	/	38.38 	&	\textbf{100.0} 	/	\textbf{100.0} 	&	\textbf{32.49} 	/	\textbf{7.04} 	&	45.02 	\\
M-ROME	&	70.27 	/	52.75 	&	63.50 	/	43.30 	&	77.29 	/	59.88 	&	26.67 	/	0.58 	&	84.58 	/	71.36 	&	78.17 	/	62.49 	&	67.99 	/	48.41 	&	26.55 	/	5.01 	&	52.43 	\\
M-MEMIT	&	99.09 	/	98.07 	&	90.07 	/	84.57 	&	98.44 	/	96.62 	&	\textbf{28.39} 	/	2.03 	&	98.34 	/	97.01 	&	91.16 	/	86.89 	&	97.91 	/	95.08 	&	27.89 	/	6.27 	&	74.86 	\\
M-PMET	&	96.41 	/	93.44 	&	85.96 	/	76.86 	&	98.35 	/	96.62 	&	27.90 	/	1.25 	&	96.74 	/	93.92 	&	88.38 	/	81.49 	&	97.99 	/	95.08 	&	27.99 	/	5.69 	&	72.75 	\\
LU-LAFNs (Ours)	&	\textbf{99.83} 	/	\textbf{99.71} 	&	\textbf{96.40} 	/	\textbf{94.41} 	&	\textbf{100.0} 	/	\textbf{100.0} 	&	26.43 	/	2.70 	&	\textbf{99.65} 	/	\textbf{99.42} 	&	\textbf{97.78} 	/	\textbf{96.43} 	&	\textbf{100.0} 	/	\textbf{100.0}	&	26.94 	/	6.17 	&	\textbf{77.87} 	\\
    \toprule
    \end{tabular}
    }
    \caption{
        The \textbf{F1/EM (\%)} results on Bi-ZsRE using Llama-3.1-8B, Qwen2-7B, and bloomz-7b1-mt as backbones.
        Results in \textbf{bold} represent the best results.
        ``\textbf{\textit{avg}}'' denotes the average value of all metrics in both two languages.
    }
    \label{table:F1-res}
\end{table*}

\subsection{Contrast Methods}

\paragraph{Fine-tuning Method.}
We directly use LoRA \cite{hu2021lora} to conduct parameter-efficient tuning for the original LLM, namely LoRA-FT.

\paragraph{MKE Method.}
ReMaKE \cite{wang2023retrievalaugmented} retrieves similar knowledge from a multilingual knowledge base as the context to instruct the model. 
Here, for the multiple languages to be edited, we retrieve\footnote{We use XLM-RoBERTa-base shared by \url{https://github.com/weixuan-wang123/ReMaKE}.} top-one question (with the answer) for each language and concatenate them as the context.

\paragraph{Adaptations of KE methods.}
We mainly adapt some Locate-then-Edit methods to MKE.
For example, ROME \cite{NEURIPS2022_6f1d43d5} modifies the output matrix of one FFN layer located following causal tracing analysis. 
MEMIT \cite{meng2023massediting} updates the output matrices of multiple FFN layers simultaneously.
PMET \cite{li2024pmet} conducts more precise editing based on MEMIT.
We extend ROME, MEMIT, and PMET to M-ROME, M-MEMIT, and M-PMET to edit multilingual knowledge simultaneously.
Specifically, since the knowledge to be edited of different languages corresponds to different answers, we train the new value for updating FFNs separately for each language.
And we estimate the previously memorized keys of FFNs for each language.

\begin{figure*}[t]
    \centering
    \includegraphics[width=\linewidth]{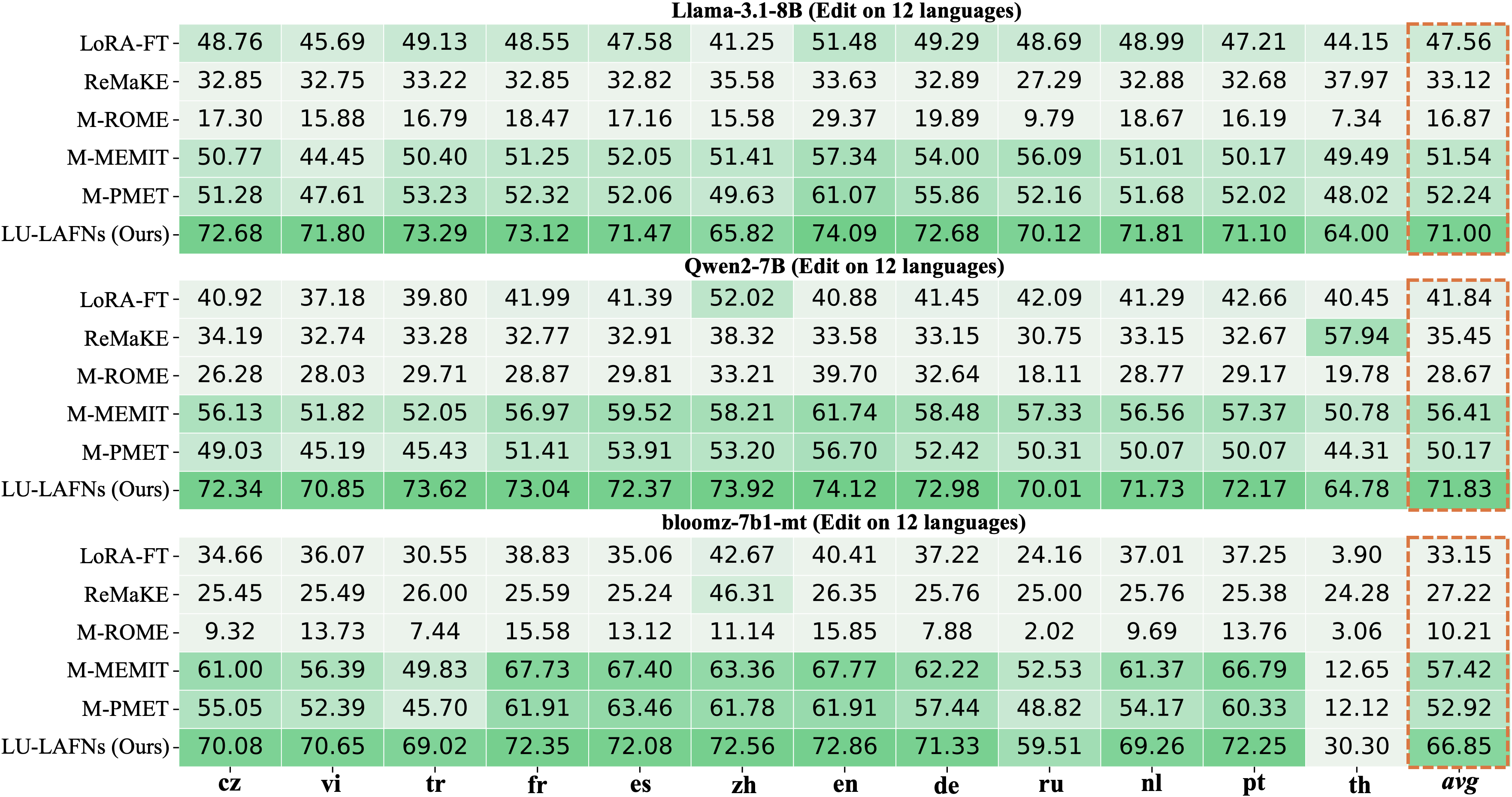}
    \caption{The average \textbf{EM (\%)} results of four metrics (Reliability, Generality, Locality, and Portability) on the MzsRE dataset using Llama-3.1-8B, Qwen2-7B, and bloomz-7b1-mt as backbones. 
    Values below 40.0 are shown in the same light color, while higher values have deeper colors indicating better performance. The orange box highlights the mean results across 12 languages.
The detailed results for each metric are listed in Appendix \ref{sec:appendix-12langs}.}
\label{fig:EM-mzsre-res}
\end{figure*}

\subsection{Experimental Results}
\paragraph{Results on Bi-ZsRE.}
Table \ref{table:F1-res} shows the F1/EM results on Bi-ZsRE using Llama-3.1-8B, Qwen2-7B, and bloomz-7b1-mt as backbones.
From the ``\textit{\textbf{avg}}'' column, the average results of all metrics show that our method outperforms other baselines significantly in all three backbones.
Particularly, our method exceeds other methods by almost >5 points in Reliability and Generality under F1\&EM. 
The superiority in Reliability indicates that updating LAFNs can edit the multilingual knowledge (needs to be edited) more effectively, and in Generality means excellent generalization on the equivalent questions that have the same semantics as the edited questions.
As for Locality, both our method and ReMaKE achieve the ``100.00'' value since the two methods do not modify the parameters of the original model during the editing process, not influencing previously learned knowledge.
However, ReMaKE performs poorly in Reliability and Generality because the retrieved examples can not instruct the model to generate correct answers. 
LoRA-FT has a good performance in Reliability among baselines (\textit{e.g.}, when Llama-3.1-8B testing on en, and when Qwen2-7B and bloomz-7b1-mt testing on zh), but it scores the lowest Locality since it dramatically modifies the original model parameters.
Additionally, the adaptions of Locate-then-Edit methods to MKE (M-ROME, M-MEMIT, and M-PMET) perform moderately among all methods.
Specifically, M-ROME is less effective than M-MEMIT and M-PMET because it only updates a single layer.
M-PMET performs the second best on Llama-3.1-8B, and M-MEMIT performs the second best on Qwen2-7B and bloomz-7b1-mt, while both are inferior to our method.
It demonstrates that the simple adaptations of these methods to MKE are less effective due to overlooking the connections of multilingual knowledge.

Portability, as a more difficult metric, measures whether the edited model can reason based on the edited knowledge via a portability question \cite{yao-etal-2023-editing, sun-etal-2024-outdated}, where the relations and objects are out of the scope of the edited knowledge.
The corresponding results show that all methods underperform on this metric without significant difference, especially when all EM results are less than 10, even than 5.
We speculate that this reasoning ability is difficult to be well-measured without a reasoning process.
We believe there is substantial room for measuring and improving portability in the future.
Moreover, we observe that Llama-3.1-8B exhibits notably superior edit performance on English compared to Chinese since Llama-3.1-8B is not fine-tuned using Chinese instruction data.
We guess that the inherent language capabilities of LLMs have a crucial impact on their edit performance.

\paragraph{Results on MzsRE.}
As for more languages, the average EM results of four metrics on MzsRE are reported in Figure \ref{fig:EM-mzsre-res}. 
(Detailed EM results of each metric for the three LLMs are listed in Table \ref{table:12langs-llama}, \ref{table:12langs-qwen}, and \ref{table:12langs-bloom} of Appendix \ref{sec:appendix-12langs}.)
Figure \ref{fig:EM-mzsre-res} shows that our method (with the deeper color) performs better than other baselines to a large extent on each language under three backbones.
Among baselines, ``M-ROME'' performs worst since this method only updates one single layer, struggling to support simultaneous editing of more language knowledge.
Other methods also underperform our method and exhibit a similar trend with the performance on Bi-ZsRE.
For the edit performance of each language, most methods perform better on English than other languages under all backbones since these LLMs are primarily proficient in English 
(due to the existence of large-scale high-quality English data).
Additionally, we also observe that the edit performance in the same language family is similar since these languages have a shared vocabulary, such as the Indo-European Family (Germanic languages: $en$, $de$, and $nl$, Slavic languages: $cz$ and $ru$, Romance languages: $es$, $fr$, and $pt$).
Moreover, Llama-3.1-8B has a worse performance on $vi$, $zh$, and $th$. Qwen2-7B also performs poorly on $vi$ and $th$ than other languages, while bloomz-7b1-mt performs badly on $tr$, $ru$, and $th$.
The different edit performance of different LLMs on various languages is probably due to the language distribution of the training dataset and the linguistic characteristics of different languages.
These results further demonstrate that the inherent language capabilities of LLMs determine the edit performance in different languages.

\section{Analysis}
In \S\ref{sec:ana-conflicts}, we initially demonstrate the knowledge conflicts of other baselines.
Then we explore the key factors affecting edit performance in \S\ref{sec:ana-settings}.
Subsequently, we compare different locating strategies to prove that using paraphrases during the locating stage can improve the edit performance (\S\ref{sec:ana-locating}).

\begin{table}[t]
    \centering
    \resizebox*{\linewidth}{!}{
    \begin{tabular}{c|cc|ll}
    \bottomrule

        \textbf{Edit Languages} &  \multicolumn{2}{c|}{\textbf{Monolingual}}	&	\multicolumn{2}{c}{\textbf{Multilingual}} \\
        
        \hline
        \textbf{Test Languages} &  \textbf{en} &	\textbf{zh} &	\makecell[c]{\textbf{en}}  &	\makecell[c]{\textbf{zh}} \\
    
    \toprule
    \multicolumn{5}{c}{\textbf{Qwen2-7B}} \\
    \bottomrule
LoRA-FT	&	40.60  &	48.60 	&	40.31 \color{Red}{($-$0.29})		&	47.37 \color{Red}{($-$1.23})	\\
ReMaKE	&	33.87 &	39.03 	&	33.78 \color{Red}{($-$0.10})		&	40.38 \color{ForestGreen}{($+$\textbf{1.35}})	\\
M-ROME	&	67.60 &	67.24  	&	53.95 \color{Red}{($-$\textbf{13.65}})		&	57.19 \color{Red}{($-$\textbf{10.05}})	\\
M-MEMIT	&	69.12  &	71.29 	&	66.69 \color{Red}{($-$2.43})		&	69.14 \color{Red}{($-$2.15})	\\
M-PMET	&	61.31  &	61.65 	&	57.77 \color{Red}{($-$3.54})		&	59.71 \color{Red}{($-$1.93})	\\
LU-LAFNs (Ours)	&	\textbf{72.35} &	\textbf{74.04} 	&	\textbf{73.39} \color{ForestGreen}{($+$\textbf{1.04}})		&	\textbf{74.47} \color{ForestGreen}{($+${0.43}})	\\
    
    \toprule
    \multicolumn{5}{c}{\textbf{bloomz-7b1-mt}} \\
    \bottomrule
LoRA-FT	&	33.83  &	40.65 	&	33.80 \color{Red}{($-$0.03})		&	40.24 \color{Red}{($-$0.41})	\\
ReMaKE	&	28.40  &	48.19 	&	25.80 \color{Red}{($-$2.60})		&	45.85 \color{Red}{($-$2.34})	\\
M-ROME	&	57.47  &	59.19 	&	39.13 \color{Red}{($-$\textbf{18.35}})		&	46.82 \color{Red}{($-$\textbf{12.37}})	\\
M-MEMIT	&	71.75  &	72.93 		&	70.32 \color{Red}{($-$1.43})	&	71.31 \color{Red}{($-$1.61})	\\
M-PMET	&	69.67  &	70.47 	&	67.04 \color{Red}{($-$2.63})		&	69.05 \color{Red}{($-$1.43})	\\
LU-LAFNs (Ours)	&	\textbf{72.81}  &	\textbf{75.34 }		&	\textbf{74.21} \color{ForestGreen}{($+$\textbf{1.40}}) 	&	\textbf{75.51} \color{ForestGreen}{($+$\textbf{0.17}})	\\
    
    \toprule
    \end{tabular}
    }
    \caption{
        The average EM results of four metrics on Bi-ZsRE using Qwen2-7B and bloomz-7b1-mt. 
        ``Monolingual'' means editing and testing on the same one language. ``Multilingual'' means editing on both $en$ and $zh$, and testing on each language, respectively.
        The values in {\color{Red}red} represent the performance decline compared with monolingual KE due to knowledge conflicts.
        The others in {\color{ForestGreen}green} represent no such conflicts.
    }
    \label{table:analysis-conflicts}
\end{table}

\subsection{Conflicts of Editing Multilingual Knowledge}\label{sec:ana-conflicts}
We conduct monolingual editing and multilingual editing experiments on Bi-ZsRE, and the results are reported in Table \ref{table:introduction} and \ref{table:analysis-conflicts}.
Referring to the red values, we can find that most methods (\textit{e.g.}, LoRA-FT, M-ROME, M-MEMIT, and M-PMET) on the three LLMs lead to conflicts when conducting MKE, resulting in the degradation of edit performance compared to monolingual KE. 
Among them, M-ROME has a dramatic decline due to the limited edit region.
By contrast, our method conducts MKE by locating and updating LAFNs, which does not cause conflicts and further improves the edit performance than monolingual KE.
Additionally, although the ReMaKE method does not cause conflicts when Qwen2-7B testing on $zh$, its edit performance is much lower than our method.
In summary, our method avoids the conflicts in MKE by locating and updating LAFNs, which represent the connections between multilingual knowledge.

\begin{table}[t]
    \centering
    \resizebox*{\linewidth}{!}{
    \begin{tabular}{c|c|c|c}
    \bottomrule
        \textbf{A Single Layer} &  \textbf{\textit{avg}} &	\textbf{ \quad Multiple Layers \quad}  &	\textbf{\textit{avg}}\\
        \hline
        0 & 66.60 & 11-12 &  74.37  \\
        5 & 72.86 & 12-13 &  74.28  \\
        10 & 72.73 & 12-31 & 74.31 \\
        \textbf{12} & \textbf{73.35} & 11-12-13 &  74.45 \\
        27 & 71.35 & \textbf{11-12-13-31} & \textbf{74.91} \\
        31 & 33.43 & 12-13-14-15-31 & 74.54 \\
    \toprule
    
    \end{tabular}
    }
    \caption{
        The average F1/EM results of different layer settings on Bi-ZsRE using Llama-3.1-8B as the backbone (when $\beta = 0.1$).
    }
    \label{table:dif-layers-res}
\end{table}

\subsection{Key Factors to Edit Performance}\label{sec:ana-settings}
In this section, we explore the key factors affecting edit performance based on Llama-3.1-8B.

\paragraph{Updated Layers of LAFNs.}
Figure \ref{fig:layer} in \S\ref{sec:3-1} has shown that the LAFNs are mostly located in some middle FFN layers and the last FFN layer.
Thus, we further evaluate our method when updating LAFNs in different layers according to the distribution of LAFNs, including updating a single layer and multiple layers (the threshold $\beta$ in Eq.(\ref{eq:n2}) is set to 0.1 for this evaluation).
The corresponding results reported in Table \ref{table:dif-layers-res} show that in the single-layer setting, the edit performance achieves the best in the 12-th layer (which has the most LAFNs in the middle layers) and worst in the last layer. 
Although the last layer also has numerous LAFNs, we conjecture that these neurons are directly related to the final outputs, and thus a single update vector is difficult to fulfill answers in all languages. 
Moreover, we find that simultaneously editing multiple layers around the 12-th layer can further improve edit performance, with the best performance observed in (11, 12, 13, 31) layers.

\paragraph{Number of LAFNs in Updated Layers.}
We also explore the influence of the threshold $\beta$ in Eq.(\ref{eq:n2}), which controls the number of LAFNs in each layer, when editing the (11, 12, 13, 31) layers.
The results in Table \ref{table:dif-threshold-res} show that when $0 \leq \beta \leq 0.4$, the edit performance does not change obviously, and the best performance is achieved when $\beta = 0.1$, that is, 67.9\% of neurons are located and modified in each layer. 
Moreover, when $\beta = 0.7$ (only updates 5.0\% LAFNs for each layer), the performance (70.25) still exceeds the baselines in Table \ref{table:F1-res} (the best is 67.40 by M-PMET), proving the effectiveness of updating LAFNs.
In summary, \textbf{both the updated layers and the number of LAFNs affect the edit performance, with the layers having a greater impact}.
The discussions of Qwen2-7B and bloomz-7b1-mt are listed in Appendix \ref{sec:appendix-settings}, which draw similar conclusions with Llama-3.1-8B.

\begin{table}[t]
    \centering
    \resizebox*{\linewidth}{!}{
    \begin{tabular}{ccc|ccc}
    \bottomrule
        \textbf{$\beta$} & \textbf{Num (Proportion)} & \textbf{\textit{avg}} &	\textbf{$\beta$} & \textbf{Num (Proportion)} &	\textbf{\textit{avg}}\\
    \hline
        0 & 14046 (98.0\%) & 74.85  & 0.5 & 1933 (13.5\%) & 73.26  \\
        \textbf{0.1} & \textbf{9738 (67.9\%)} & \textbf{74.91} & 0.6 & 1195 (8.3\%) &  72.08  \\
        0.2 & 6729 (46.9\%) & 74.79 & 0.7 & 720 (5.0\%) & 70.25 \\
        0.3 & 4613 (32.2\%) & 74.74 & 0.8 &  418 (2.9\%) & 66.30 \\
        0.4 & 3045 (21.2\%) & 74.44 & 0.9 & 223 (1.6\%) & 50.56 \\
    \toprule
    
    \end{tabular}
    }
    \caption{
        The average F1/EM results of different $\beta$ on Bi-ZsRE using Llama-3.1-8B as the backbone when editing (11, 12, 13, 31) layers. The ``Num  (Proportion)'' represents the average number and proportion of LAFNs on each updated layer.
    }
    \label{table:dif-threshold-res}
\end{table}

\begin{table}[t]
    \centering
    \resizebox*{\linewidth}{!}{
    \begin{tabular}{c|ccc}
    \bottomrule
        \textbf{Methods} &  \textbf{Llama-3.1-8B} &	\textbf{Qwen2-7B}  &	 \textbf{bloomz-7b1-mt}\\
    \hline
        LU-LAFNs (Ours) &  \textbf{74.91}   &  \textbf{77.74}  &  \textbf{77.87}  \\  
         No-PGs &  74.75 ($\downarrow$ 0.16)   &  77.69 ($\downarrow$ 0.05)  &   77.70 ($\downarrow$ 0.17) \\ 
        All &   74.85 ($\downarrow$ 0.06)  &  77.71 ($\downarrow$ 0.03)  &  77.13 ($\downarrow$ \textbf{0.74}) \\  
         Random &  74.69 ($\downarrow$ \textbf{0.22})  &  77.61 ($\downarrow$ \textbf{0.13})  &  77.66 ($\downarrow$ 0.21) \\  
        
 
    \toprule
    \end{tabular}
    }
    \caption{
        The average F1/EM results of different locating strategies on Bi-zsRE using Llama-3.1-8B, Qwen2-7B, and bloomz-7b1-mt as backbones. 
    }
    \label{table:ablation-res}
\end{table}

\subsection{Different Locating Strategies}\label{sec:ana-locating}
To verify the effectiveness of using paraphrases during the locating stage, we compare three different locating strategies with the original LU-LAFNs: 
(1) \textbf{No-PGs}: not using paraphrases to assist in locating LAFNs, \textit{i.e.}, only using a single sentence in each language;
(2) \textbf{All}: modifying all neurons of the same layers as LU-LAFNs without locating the set of LAFNs; 
(3) \textbf{Random}: randomly selecting the same number of neurons in the same layers to modify.
The results in Table \ref{table:ablation-res} show that the performance of all these three settings declines compared to the proposed method. 
These results demonstrate that using paraphrases during the locating stage can improve the edit performance since it can locate the LAFNs that are more semantically relevant to the multilingual knowledge to be edited.

\section{Conclusion}
In this work, we first identify language-agnostic factual neurons (LAFNs) in LLMs that represent the factual knowledge shared across different languages and imply semantic connections between multilingual knowledge.
Then, we propose a new method LU-LAFNs to conduct MKE by locating and updating LAFNs.
The experimental results demonstrate our method avoids knowledge conflicts and achieves the best MKE performance.

\section*{Limitations}
In our approach, it is necessary to provide the aligned multilingual knowledge to be edited and their corresponding multilingual subjects, which is directly available in both Bi-ZsRE and MzsRE datasets.
However, for other datasets that do not contain this information, we first need to preprocess the data to support our method.
For example, if there is no corresponding multilingual data available, using translation API can translate the existing knowledge to be edited to other languages. 
If the corresponding subjects are not annotated, existing LLMs can be utilized to identify the aligned multilingual subjects in the sentences of each language. 
These preprocessing steps can be easily implemented by calling existing tools.
Moreover, the current method for determining whether a subject exists in the cache adopts the exact-match approach, which is too strict. 
We will optimize it to a fuzzy matching method in future work to enhance the performance in practical application scenarios. 

Furthermore, our method performs poorly in the Portability metric, which measures whether the edited model can reason based on the edited knowledge. 
Recently, \citeauthor{khandelwal2024crosslingualmultihopknowledgeediting} propose the cross-lingual multi-hop knowledge editing benchmark CROLIN-MQUAKE based on MQUAKE \cite{zhong2023mquakeassessingknowledgeediting} to test the multi-hop reasoning ability of the edited model. 
Next, we will test our method on this benchmark and further improve our method in reasoning scenarios.

\section*{Acknowledgments}
The research work described in this paper has been supported by the National Nature Science Foundation of China (No. 62476023, 61976016, 62376019, 61976015), and the authors would like to thank the anonymous reviewers for their valuable comments and suggestions to improve this paper.

\bibliography{custom}

\appendix

\section{Detailed Experimental Settings}
\subsection{Evaluation Metrics}\label{sec:appendix-metrics}
The details of the four metrics are as follows \cite{wang2023crosslingual}.
\textbf{Reliability} measures the average accuracy on the edit case. When receiving $x_e$ as input, the edited model $\mathcal{F}_{\theta}^{'}$ should output $y_e$.
\textbf{Generality} evaluates the average accuracy on the equivalent cases as the edit case. For instance, when receiving a rephrased sentence of $x_e$, the edited model $\mathcal{F}_{\theta}^{'}$ is also expected to output $y_e$.
\textbf{Locality} assesses the accuracy of the edited model on the irrelevant samples.
When the input $x$ is irrelevant with $x_e$, $\mathcal{F}_{\theta}^{'}(x)$ should be the same as $\mathcal{F}_{\theta}(x)$ ideally.
\textbf{Portability} measures the robust generalization of the edited model via a portability question that needs reasoning based on the edited knowledge. When receiving the portability question as input, the edited model $\mathcal{F}_{\theta}^{'}$ is expected to output the golden answer to demonstrate the model indeed learns the knowledge.

\subsection{Supported Languages of LLMs}\label{sec:appendix-langsfor3}
Llama-3.1-8B is fine-tuned with high-quality multilingual instruction data including English, French, German, Portuguese, Spanish, and Thai, and also has a certain degree of generalization ability to the other 6 languages in MzsRE.
Qwen2-7B supports all 12 languages\footnote{\url{https://qwenlm.github.io/zh/blog/qwen2/}} in MzsRE.
The bloomz-7b1-mt model is finetuned on the cross-lingual task mixture (xP3mt\footnote{\url{https://huggingface.co/datasets/bigscience/xP3mt}}) across 46 languages and 16 NLP tasks and has the capability of cross-lingual generalization to unseen tasks and languages.

\subsection{The Instruction for generating paraphrases} \label{sec:appendix-para-instuction}
We call the Qwen2-72B-instruct API (from the ALIYUN platform) to generate the paraphrase set $P_\ell$ for more precisely locating neurons.
We directly use the default generation configs.
The English version of the instruction for inputting Qwen2-72B-instruct is ``You are an expert at sentence rewriting. Below I will give you a subject and a question containing the subject. Please give me 30 questions including this subject in English. They must have the same semantics as the given question. 
Subject: \{\}.
Question containing this Subject: \{\}''.

\section{Cross-Lingual Experiments}\label{sec:appendix-cross-lingual}
In the initial stage, we conduct cross-lingual experiments on Llama2-7B \cite{touvron2023llama}, i.e., we only utilize monolingual knowledge to edit the model and then test the edit performance on other languages. 
The results in Table \ref{table:cross-lingual} show that our method has better generalization on unseen languages than ROME/MEMIT/PEMT. 
However, there is still a large gap between the editing performance on unedited languages and that on edited languages.
Therefore, we mainly focus on multilingual knowledge editing in this paper, which performs better in updating multilingual knowledge simultaneously.

\begin{table}[t]
    \centering
    \resizebox*{\linewidth}{!}{
    \begin{tabular}{c|cc|cc}
    \bottomrule

        \textbf{Edit on} &  \textbf{en} & \textbf{en}	&	\makecell[c]{\textbf{zh}}  &	\makecell[c]{\textbf{zh}} \\
        \hline
        \textbf{Test on} &  \textbf{en} &	\textbf{zh} &	\makecell[c]{\textbf{en}}  &	\makecell[c]{\textbf{zh}} \\
        
        \hline
        
ROME	&	72.96 &	 35.11 	&	41.46 		&	 47.61 	\\
MEMIT	&	76.26  &	36.56   	&		42.64 	&	48.41  	\\
PEMT	&	77.18  &	36.00   	&	42.69 		&	48.12  	\\
LU-LAFNs (Ours)	&	\textbf{79.96}  &	\textbf{37.34}   	&	\textbf{43.74} 		&	 \textbf{55.57} 	\\
    
    \toprule
    \end{tabular}
    }
    \caption{
        The average F1 results of four metrics on Bi-ZsRE using Llama2-7B as the backbone under the cross-lingual edit setting. 
    }
    \label{table:cross-lingual}
\end{table}

\section{Detailed Results on MzsRE}\label{sec:appendix-12langs}
The detailed EM results of four metrics on MzsRE for Llama-3.1-8B, Qwen2-7B, and bloomz-7b1-mt are listed in Table  \ref{table:12langs-llama}, \ref{table:12langs-qwen}, and \ref{table:12langs-bloom}, respectively.

\section{Different Settings of Qwen2-7B and bloomz-7b1-mt}\label{sec:appendix-settings}
We report the edit performance of Qwen2-7B and bloomz-7b1-mt under different layers (Table \ref{table:dif-layers-res2}) and different values of $\beta$ (Table \ref{table:dif-threshold-res2}).
The changes in edit performance under different settings are similar to Llama-3.1-8B.
Qwen2-7B achieves the best result when editing (19, 20, 21, 27) layers and $\beta = 0$, and bloomz-7b1-mt performs best when editing (9, 10, 11, 29) layers and $\beta = 0.2$.
Additionally, when $\beta = 0.5$ on editing Qwen2-7B (only updates 5.1\% LAFNs for each layer), the result (75.44) exceeds all baselines in Table \ref{table:F1-res} (the best result is 73.71 of M-MEMIT). 
And when $\beta = 0.5$ on editing bloomz-7b1-mt (only updates 1.9\% LAFNs for each layer), the result (77.41) exceeds all baselines in Table \ref{table:F1-res} (the best result is 74.86 of M-MEMIT).
Furthermore, the appropriate layer setting is more crucial to edit performance than the threshold $\beta$.

\begin{table}[t]
    \centering
    \resizebox*{\linewidth}{!}{
    \begin{tabular}{c|c|c|c}
    \bottomrule
        \textbf{A Single Layer} &  \textbf{\textit{avg}} &	\textbf{ \quad \quad  Multiple Layers \quad \quad }  &	\textbf{\textit{avg}}\\
    \toprule
    \multicolumn{4}{c}{\textbf{Qwen2-7B}} \\
    \bottomrule
        0 & 61.16 & 19-20 &  77.49 \\
        5 & 34.35 & 20-21 &  77.29  \\
        10 & 76.02 & 19-20-21 & 77.51 \\
        18 & 76.33 & 18-19-20-21 &  77.36 \\
        \textbf{20} & \textbf{76.56} & \textbf{19-20-21-27} & \textbf{77.53} \\
        27 & 33.43 &  18-19-20-21-27  & 77.51 \\
    \toprule
    \multicolumn{4}{c}{\textbf{bloomz-7b1-mt}} \\
    \bottomrule
        0 & 48.18 & 10-11 &  77.80 \\
        2 & 75.31 & 10-15 &  77.79  \\
        4 & 71.63 & 9-10-11 & 77.83 \\
        \textbf{10} & \textbf{77.54} & 14-15-16 &  77.83 \\
        15 & 77.49 & \textbf{9-10-11-29} & \textbf{77.84} \\
        29 & 31.25 &  14-15-16-29 & 77.65 \\
    \toprule
    
    \end{tabular}
    }
    \caption{
        The results of different layer settings on Bi-ZsRE using Qwen2-7B and bloomz-7b1-mt as backbones (when $\beta = 0.1$).
    }
    \label{table:dif-layers-res2}
\end{table}

\begin{table}[t]
    \centering
    \resizebox*{\linewidth}{!}{
    \begin{tabular}{ccc|ccc}
    \bottomrule
        \textbf{$\beta$} & \textbf{Num (Proportion)} & \textbf{\textit{avg}} &	\textbf{$\beta$} & \textbf{Num (Proportion)} &	\textbf{\textit{avg}}\\
    \toprule
    \multicolumn{6}{c}{\textbf{Qwen2-7B, layers=19-20-21-27}} \\
    \bottomrule
        \textbf{0} & \textbf{18190 (96.0\%)} & \textbf{77.74}  & 0.5 & 962 (5.1\%) &  75.44 \\
        0.1 & 8907 (47.0\%) & 77.53 & 0.6 & 603 (3.2\%) &  71.84  \\
        0.2 & 4421 (23.3\%) & 77.25 & 0.7 & 375 (2.0\%) & 54.88 \\
        0.3 & 2539 (13.4\%) & 76.69 & 0.8 & 226 (1.2\%)  & 38.04 \\
        0.4 & 1543 (8.1\%) & 76.29 & 0.9 & 125 (0.7\%) & 34.35 \\
    \toprule
    \multicolumn{6}{c}{\textbf{bloomz-7b1-mt, layers=9-10-11-29}} \\
    \bottomrule
        0 & 15220 (92.9\%) &  77.36 & 0.5 & 316 (1.9\%) &  77.41 \\
        0.1 & 8169 (49.9\%) & 77.84 & 0.6 & 114 (0.7\%) &  45.50  \\
        \textbf{0.2} & \textbf{4201 (25.6\%)} & \textbf{77.87} & 0.7 & 43 (0.3\%) & 33.03 \\
        0.3 & 2009 (12.3\%) & 77.68 & 0.8 & 17 (0.1\%)  & 31.58 \\
        0.4 & 844 (5.2\%) & 77.21 & 0.9 & 0 (0.0\%) & 0.00 \\
    \toprule
    
    \end{tabular}
    }
    \caption{
        The results of different $\beta$  on Bi-ZsRE using Qwen2-7B and bloomz-7b1-mt as backbones under the best layer setting.
        The ``Num  (Proportion)'' represents the average number and proportion of LAFNs on each updated layer.
    }
    \label{table:dif-threshold-res2}
\end{table}

\begin{table*}[t] 
    \centering
    \resizebox*{\linewidth}{!}{
    \begin{tabular}{l|ccccccccccccc}
    \bottomrule
     \textbf{Methods} & \textbf{cz}     & \textbf{vi}     & \textbf{tr}     & \textbf{fr}     & \textbf{es}     & \textbf{zh}     & \textbf{en}     & \textbf{de}     & \textbf{ru}     & \textbf{nl}     & \textbf{pt}     & \textbf{th}     & \textbf{\textit{avg}}\\
    \toprule
    \multicolumn{14}{c}{\textbf{Reliability}} \\
    \bottomrule
    LoRA-FT	&	98.65 	&	96.10 	&	98.79 	&	97.17 	&	96.77 	&	81.29 	&	99.19 	&	98.38 	&	98.25 	&	99.19 	&	95.96 	&	96.23 	&	\textbf{96.33} 	\\
ReMaKE	&	15.36 	&	15.23 	&	16.17 	&	15.36 	&	15.36 	&	19.95 	&	16.85 	&	15.50 	&	3.10 	&	15.50 	&	15.09 	&	27.76 	&	15.94 	\\
M-ROME	&	31.36 	&	26.24 	&	29.61 	&	31.90 	&	30.28 	&	25.71 	&	48.18 	&	35.13 	&	15.75 	&	32.97 	&	27.05 	&	13.73 	&	28.99 	\\
M-MEMIT	&	69.85 	&	54.24 	&	66.49 	&	67.43 	&	66.49 	&	63.53 	&	74.43 	&	71.06 	&	84.12 	&	69.58 	&	67.70 	&	82.91 	&	69.82 	\\
M-PMET	&	68.10 	&	60.16 	&	70.52 	&	66.62 	&	65.95 	&	61.64 	&	80.48 	&	73.22 	&	75.24 	&	67.43 	&	68.10 	&	74.83 	&	69.36 	\\
LU-LAFNs (Ours)	&	97.98 	&	96.90 	&	97.17 	&	97.58 	&	95.29 	&	81.83 	&	98.12 	&	96.77 	&	94.08 	&	95.83 	&	94.35 	&	88.29 	&	94.52 	\\
    \toprule
    \multicolumn{14}{c}{\textbf{Generality}} \\
    \bottomrule
    LoRA-FT	&	88.83 	&	81.83 	&	89.10 	&	89.77 	&	88.69 	&	76.99 	&	94.89 	&	91.52 	&	88.16 	&	91.66 	&	87.21 	&	78.20 	&	87.24 	\\
ReMaKE	&	15.50 	&	15.36 	&	16.17 	&	15.36 	&	15.36 	&	21.56 	&	16.85 	&	15.50 	&	6.33 	&	15.63 	&	14.96 	&	23.72 	&	16.03 	\\
M-ROME	&	29.07 	&	25.17 	&	28.26 	&	30.15 	&	28.26 	&	24.36 	&	48.32 	&	35.53 	&	15.07 	&	31.22 	&	25.98 	&	9.42 	&	27.57 	\\
M-MEMIT	&	59.35 	&	46.03 	&	58.68 	&	57.74 	&	58.14 	&	56.53 	&	63.93 	&	61.10 	&	69.04 	&	58.55 	&	57.60 	&	44.41 	&	57.59 	\\
M-PMET	&	61.91 	&	54.24 	&	66.22 	&	62.31 	&	62.45 	&	57.20 	&	75.24 	&	68.64 	&	64.87 	&	62.31 	&	62.72 	&	45.76 	&	61.99 	\\
LU-LAFNs (Ours)	&	91.12 	&	88.69 	&	92.87 	&	93.41 	&	88.69 	&	77.93 	&	95.96 	&	91.92 	&	84.39 	&	89.77 	&	88.16 	&	66.35 	&	\textbf{87.44} 	\\
    \toprule
    \multicolumn{14}{c}{\textbf{Locality}} \\
    \bottomrule
    LoRA-FT	&	3.77 	&	2.42 	&	4.31 	&	3.63 	&	2.42 	&	3.36 	&	5.92 	&	3.63 	&	4.17 	&	2.56 	&	2.83 	&	1.08 	&	3.34 	\\
ReMaKE	&	100.00 	&	100.00 	&	100.00 	&	100.00 	&	100.00 	&	100.00 	&	100.00 	&	100.00 	&	99.73 	&	100.00 	&	100.00 	&	100.00 	&	99.98 	\\
M-ROME	&	7.54 	&	10.50 	&	7.00 	&	10.63 	&	8.61 	&	10.50 	&	17.09 	&	7.40 	&	6.59 	&	9.29 	&	9.83 	&	6.06 	&	9.25 	\\
M-MEMIT	&	71.74 	&	74.70 	&	72.54 	&	76.72 	&	80.75 	&	82.23 	&	85.87 	&	80.48 	&	68.51 	&	73.49 	&	71.87 	&	70.52 	&	75.79 	\\
M-PMET	&	72.95 	&	72.68 	&	72.01 	&	77.25 	&	76.99 	&	76.58 	&	83.98 	&	78.06 	&	66.49 	&	74.02 	&	73.76 	&	70.93 	&	74.64 	\\
LU-LAFNs (Ours)	&	100.00 	&	100.00 	&	100.00 	&	100.00 	&	100.00 	&	100.00 	&	100.00 	&	100.00 	&	99.73 	&	100.00 	&	100.00 	&	100.00 	&	\textbf{99.98} 	\\
    \toprule
    \multicolumn{14}{c}{\textbf{Portability}} \\
    \bottomrule
    LoRA-FT	&	3.77 	&	2.42 	&	4.31 	&	3.63 	&	2.42 	&	3.36 	&	5.92 	&	3.63 	&	4.17 	&	2.56 	&	2.83 	&	1.08 	&	\textbf{3.34} 	\\
ReMaKE	&	0.54 	&	0.40 	&	0.54 	&	0.67 	&	0.54 	&	0.81 	&	0.81 	&	0.54 	&	0.00 	&	0.40 	&	0.67 	&	0.40 	&	0.53 	\\
M-ROME	&	1.21 	&	1.62 	&	2.29 	&	1.21 	&	1.48 	&	1.75 	&	3.90 	&	1.48 	&	1.75 	&	1.21 	&	1.88 	&	0.13 	&	1.66 	\\
M-MEMIT	&	2.15 	&	2.83 	&	3.90 	&	3.10 	&	2.83 	&	3.36 	&	5.11 	&	3.36 	&	2.69 	&	2.42 	&	3.50 	&	0.13 	&	2.95 	\\
M-PMET	&	2.15 	&	3.36 	&	4.17 	&	3.10 	&	2.83 	&	3.10 	&	4.58 	&	3.50 	&	2.02 	&	2.96 	&	3.50 	&	0.54 	&	2.98 	\\
LU-LAFNs (Ours)	&	1.62 	&	1.62 	&	3.10 	&	1.48 	&	1.88 	&	3.50 	&	2.29 	&	2.02 	&	2.29 	&	1.62 	&	1.88 	&	1.35 	&	2.05 	\\

    \toprule             
    \end{tabular}
    }
    \caption{
        The \textbf{EM} results on the MzsRE dataset using \textbf{Llama-3.1-8B} as the backbone.
    }
    \label{table:12langs-llama}
\end{table*}

\begin{table*}[t] 
    \centering
    \resizebox*{\linewidth}{!}{
    \begin{tabular}{l|ccccccccccccc}
    \bottomrule
     \textbf{Methods} & \textbf{cz}     & \textbf{vi}     & \textbf{tr}     & \textbf{fr}     & \textbf{es}     & \textbf{zh}     & \textbf{en}     & \textbf{de}     & \textbf{ru}     & \textbf{nl}     & \textbf{pt}     & \textbf{th}     & \textbf{\textit{avg}}\\
    \toprule
    \multicolumn{14}{c}{\textbf{Reliability}} \\
    \bottomrule
    LoRA-FT	&	85.60 	&	78.73 	&	83.04 	&	83.98 	&	85.60 	&	94.08 	&	80.62 	&	83.18 	&	90.04 	&	85.60 	&	88.69 	&	91.66 	&	85.90 	\\
ReMaKE	&	18.03 	&	15.21 	&	16.29 	&	15.21 	&	15.61 	&	23.55 	&	16.82 	&	16.02 	&	9.42 	&	16.15 	&	15.07 	&	60.57 	&	19.83 	\\
M-ROME	&	33.51 	&	32.97 	&	38.49 	&	32.84 	&	31.90 	&	34.32 	&	47.78 	&	38.76 	&	17.23 	&	35.53 	&	30.69 	&	26.51 	&	33.38 	\\
M-MEMIT	&	86.00 	&	74.83 	&	73.62 	&	83.31 	&	84.66 	&	77.25 	&	89.37 	&	88.29 	&	87.21 	&	84.66 	&	82.23 	&	88.16 	&	83.30 	\\
M-PMET	&	68.91 	&	59.08 	&	59.89 	&	70.52 	&	72.14 	&	68.24 	&	77.12 	&	72.81 	&	72.41 	&	68.24 	&	67.43 	&	70.79 	&	68.97 	\\
LU-LAFNs (Ours)	&	97.84 	&	96.76 	&	98.11 	&	98.92 	&	96.49 	&	98.92 	&	99.46 	&	98.52 	&	94.60 	&	97.44 	&	97.44 	&	91.36 	&	\textbf{97.16} 	\\
    \toprule
    \multicolumn{14}{c}{\textbf{Generality}} \\
    \bottomrule
    LoRA-FT	&	72.41 	&	63.53 	&	70.12 	&	76.31 	&	73.62 	&	84.79 	&	74.02 	&	75.50 	&	73.62 	&	74.56 	&	77.25 	&	65.14 	&	73.41 	\\
ReMaKE	&	18.03 	&	15.48 	&	16.55 	&	15.61 	&	15.88 	&	26.51 	&	16.82 	&	16.15 	&	13.59 	&	16.29 	&	15.34 	&	68.37 	&	21.22 	\\
M-ROME	&	32.44 	&	29.61 	&	34.32 	&	30.28 	&	28.40 	&	30.82 	&	42.93 	&	36.20 	&	14.54 	&	32.03 	&	30.01 	&	15.07 	&	29.72 	\\
M-MEMIT	&	71.47 	&	61.91 	&	67.03 	&	71.47 	&	75.37 	&	69.18 	&	77.66 	&	73.49 	&	74.29 	&	69.58 	&	70.52 	&	51.41 	&	69.45 	\\
M-PMET	&	59.08 	&	47.91 	&	52.36 	&	62.05 	&	64.20 	&	58.14 	&	67.83 	&	61.91 	&	60.57 	&	58.28 	&	55.72 	&	40.65 	&	57.39 	\\
LU-LAFNs (Ours)	&	89.88 	&	85.02 	&	93.66 	&	91.50 	&	90.82 	&	91.77 	&	94.74 	&	91.23 	&	83.54 	&	88.39 	&	88.93 	&	66.80 	&	\textbf{88.02} 	\\
    \toprule
    \multicolumn{14}{c}{\textbf{Locality}} \\
    \bottomrule
    LoRA-FT	&	3.10 	&	4.71 	&	3.90 	&	4.98 	&	3.77 	&	21.53 	&	3.90 	&	4.44 	&	1.88 	&	3.10 	&	2.29 	&	3.63 	&	5.10 	\\
ReMaKE	&	99.87 	&	99.87 	&	99.87 	&	99.87 	&	99.87 	&	100.00 	&	100.00 	&	99.87 	&	99.73 	&	99.87 	&	99.87 	&	99.73 	&	99.87 	\\
M-ROME	&	37.95 	&	48.45 	&	44.55 	&	51.55 	&	58.14 	&	65.55 	&	66.22 	&	54.51 	&	39.84 	&	46.57 	&	55.32 	&	37.42 	&	50.51 	\\
M-MEMIT	&	65.28 	&	68.78 	&	63.93 	&	70.79 	&	75.91 	&	82.37 	&	76.58 	&	69.72 	&	65.14 	&	69.99 	&	74.29 	&	63.26 	&	70.50 	\\
M-PMET	&	66.49 	&	72.27 	&	67.16 	&	71.60 	&	76.99 	&	83.31 	&	79.14 	&	73.22 	&	66.22 	&	71.74 	&	74.97 	&	65.41 	&	72.38 	\\
LU-LAFNs (Ours)	&	100.00 	&	100.00 	&	100.00 	&	100.00 	&	100.00 	&	100.00 	&	100.00 	&	100.00 	&	100.00 	&	100.00 	&	100.00 	&	100.00 	&	\textbf{100.00} 	\\
    \toprule
    \multicolumn{14}{c}{\textbf{Portability}} \\
    \bottomrule
    LoRA-FT	&	2.56 	&	1.75 	&	2.15 	&	2.69 	&	2.56 	&	7.67 	&	4.98 	&	2.69 	&	2.83 	&	1.88 	&	2.42 	&	1.35 	&	\textbf{2.96} 	\\
ReMaKE	&	0.81 	&	0.40 	&	0.40 	&	0.40 	&	0.27 	&	3.23 	&	0.67 	&	0.54 	&	0.27 	&	0.27 	&	0.40 	&	3.10 	&	0.90 	\\
M-ROME	&	1.21 	&	1.08 	&	1.48 	&	0.81 	&	0.81 	&	2.15 	&	1.88 	&	1.08 	&	0.81 	&	0.94 	&	0.67 	&	0.13 	&	1.09 	\\
M-MEMIT	&	1.75 	&	1.75 	&	3.63 	&	2.29 	&	2.15 	&	4.04 	&	3.36 	&	2.42 	&	2.69 	&	2.02 	&	2.42 	&	0.27 	&	2.40 	\\
M-PMET	&	1.62 	&	1.48 	&	2.29 	&	1.48 	&	2.29 	&	3.10 	&	2.69 	&	1.75 	&	2.02 	&	2.02 	&	2.15 	&	0.40 	&	1.94 	\\
LU-LAFNs (Ours)	&	1.62 	&	1.62 	&	2.70 	&	1.75 	&	2.16 	&	4.99 	&	2.29 	&	2.16 	&	1.89 	&	1.08 	&	2.29 	&	0.94 	&	2.12 	\\

    \toprule             
    \end{tabular}
    }
    \caption{
        The \textbf{EM }results on the MzsRE dataset using \textbf{Qwen2-7B} as the backbone.
    }
    \label{table:12langs-qwen}
\end{table*}

\begin{table*}[htbp] 
    \centering
    \resizebox*{\linewidth}{!}{
    \begin{tabular}{l|ccccccccccccc}
    \bottomrule
     \textbf{Methods} & \textbf{cz}     & \textbf{vi}     & \textbf{tr}     & \textbf{fr}     & \textbf{es}     & \textbf{zh}     & \textbf{en}     & \textbf{de}     & \textbf{ru}     & \textbf{nl}     & \textbf{pt}     & \textbf{th}     & \textbf{\textit{avg}}\\
    \toprule
    \multicolumn{14}{c}{\textbf{Reliability}} \\
    \bottomrule
    LoRA-FT	&	78.06 	&	74.43 	&	67.97 	&	77.12 	&	71.47 	&	89.64 	&	79.14 	&	79.68 	&	59.62 	&	78.47 	&	73.89 	&	8.88 	&	69.86 	\\
ReMaKE	&	0.83 	&	0.97 	&	2.07 	&	1.10 	&	0.41 	&	39.45 	&	2.48 	&	1.66 	&	0.00 	&	1.79 	&	0.83 	&	0.00 	&	4.30 	\\
M-ROME	&	12.65 	&	21.27 	&	10.23 	&	23.42 	&	21.53 	&	17.23 	&	25.84 	&	14.54 	&	1.62 	&	15.07 	&	20.46 	&	4.44 	&	15.69 	\\
M-MEMIT	&	92.19 	&	75.64 	&	66.49 	&	95.15 	&	94.35 	&	83.71 	&	95.83 	&	93.54 	&	81.43 	&	92.46 	&	94.35 	&	14.67 	&	81.65 	\\
M-PMET	&	78.87 	&	67.29 	&	58.55 	&	83.31 	&	86.81 	&	81.97 	&	85.46 	&	84.79 	&	74.29 	&	75.37 	&	82.50 	&	13.06 	&	72.69 	\\
LU-LAFNs (Ours)	&	95.79 	&	97.69 	&	93.07 	&	98.23 	&	95.65 	&	96.74 	&	98.64 	&	97.69 	&	73.78 	&	96.60 	&	97.28 	&	13.99 	&	\textbf{87.93} 	\\
    \toprule
    \multicolumn{14}{c}{\textbf{Generality}} \\
    \bottomrule
    LoRA-FT	&	53.57 	&	60.30 	&	49.53 	&	65.14 	&	61.91 	&	73.49 	&	71.60 	&	65.81 	&	32.57 	&	63.66 	&	63.39 	&	4.85 	&	55.49 	\\
ReMaKE	&	0.97 	&	0.97 	&	1.93 	&	1.24 	&	0.55 	&	39.31 	&	2.90 	&	1.38 	&	0.00 	&	1.24 	&	0.69 	&	0.00 	&	4.27 	\\
M-ROME	&	13.32 	&	19.65 	&	10.36 	&	22.75 	&	19.65 	&	15.21 	&	24.50 	&	14.00 	&	1.48 	&	13.73 	&	19.52 	&	3.90 	&	14.84 	\\
M-MEMIT	&	75.50 	&	61.37 	&	55.45 	&	83.71 	&	84.66 	&	72.14 	&	82.50 	&	76.18 	&	68.91 	&	73.62 	&	82.37 	&	9.96 	&	68.86 	\\
M-PMET	&	63.39 	&	55.32 	&	48.18 	&	71.33 	&	75.64 	&	68.64 	&	69.18 	&	65.95 	&	60.30 	&	60.70 	&	68.51 	&	8.75 	&	59.66 	\\
LU-LAFNs (Ours)	&	82.74 	&	83.56 	&	80.57 	&	89.67 	&	90.62 	&	89.40 	&	91.30 	&	85.87 	&	63.04 	&	79.21 	&	89.40 	&	10.05 	&	\textbf{77.95} 	\\
    \toprule
    \multicolumn{14}{c}{\textbf{Locality}} \\
    \bottomrule
    LoRA-FT	&	6.06 	&	7.54 	&	3.77 	&	11.17 	&	5.38 	&	3.90 	&	8.34 	&	2.29 	&	3.23 	&	5.11 	&	9.29 	&	1.75 	&	5.65 	\\
ReMaKE	&	100.00 	&	100.00 	&	100.00 	&	100.00 	&	100.00 	&	100.00 	&	100.00 	&	100.00 	&	100.00 	&	100.00 	&	100.00 	&	97.10 	&	99.76 	\\
M-ROME	&	11.17 	&	13.86 	&	9.15 	&	16.02 	&	11.17 	&	10.77 	&	12.79 	&	2.96 	&	4.98 	&	9.69 	&	14.80 	&	3.77 	&	10.09 	\\
M-MEMIT	&	75.37 	&	87.48 	&	75.91 	&	91.12 	&	89.50 	&	91.92 	&	90.98 	&	78.20 	&	58.55 	&	78.33 	&	88.96 	&	25.98 	&	77.69 	\\
M-PMET	&	77.12 	&	86.27 	&	75.24 	&	92.33 	&	90.31 	&	91.39 	&	91.79 	&	78.20 	&	59.22 	&	79.81 	&	89.10 	&	26.65 	&	78.12 	\\
LU-LAFNs (Ours)	&	100.00 	&	100.00 	&	100.00 	&	100.00 	&	100.00 	&	100.00 	&	100.00 	&	100.00 	&	100.00 	&	100.00 	&	100.00 	&	97.15 	&	\textbf{99.76} 	\\
    \toprule
    \multicolumn{14}{c}{\textbf{Portability}} \\
    \bottomrule
    LoRA-FT	&	0.94 	&	2.02 	&	0.94 	&	1.88 	&	1.48 	&	3.63 	&	2.56 	&	1.08 	&	1.21 	&	0.81 	&	2.42 	&	0.13 	&	1.59 	\\
ReMaKE	&	0.00 	&	0.00 	&	0.00 	&	0.00 	&	0.00 	&	6.48 	&	0.00 	&	0.00 	&	0.00 	&	0.00 	&	0.00 	&	0.00 	&	0.54 	\\
M-ROME	&	0.13 	&	0.13 	&	0.00 	&	0.13 	&	0.13 	&	1.35 	&	0.27 	&	0.00 	&	0.00 	&	0.27 	&	0.27 	&	0.13 	&	0.23 	\\
M-MEMIT	&	0.94 	&	1.08 	&	1.48 	&	0.94 	&	1.08 	&	5.65 	&	1.75 	&	0.94 	&	1.21 	&	1.08 	&	1.48 	&	0.00 	&	1.47 	\\
M-PMET	&	0.81 	&	0.67 	&	0.81 	&	0.67 	&	1.08 	&	5.11 	&	1.21 	&	0.81 	&	1.48 	&	0.81 	&	1.21 	&	0.00 	&	1.22 	\\
LU-LAFNs (Ours)	&	1.77 	&	1.36 	&	2.45 	&	1.49 	&	2.04 	&	4.08 	&	1.49 	&	1.77 	&	1.22 	&	1.22 	&	2.31 	&	0.00 	&	\textbf{1.77} 	\\

    \toprule             
    \end{tabular}
    }
    \caption{
        The \textbf{EM} results on the MzsRE dataset using \textbf{bloomz-7b1-mt} as the backbone.
    }
    \label{table:12langs-bloom}
\end{table*}

\end{document}